\documentclass[]{raa}            % referee version: for submission

\usepackage{graphicx,times}             %for PS/EPS graphics inclusion, new
\usepackage{subcaption}
\usepackage{natbib}
\usepackage{amssymb,amsmath}
\usepackage{multirow}
\bibpunct{(}{)}{;}{a}{}{,}

\usepackage[pagebackref=true]{hyperref}

\begin{document}

  \title{AI Agent for Source Finding by SoFiA-2 for SKA-SDC2
}
%   \subtitle{I. Place Your Subtitle Here}

   \volnopage{Vol.0 (20xx) No.0, 000--000}      %%preserved for Editor. DOn't remove!
   \setcounter{page}{1}          %%starting page, preserved for Editor. DOn't remove!

   \author{Xingchen Zhou %(周兴晨) %% Put your Chinese name in "( )" if you like. Note to open line 11 
      \inst{1}
   \and Nan Li
      \inst{1}
   \and Peng Jia
      \inst{2}
    \and Yingfeng Liu
        \inst{1}
    \and Furen Deng
        \inst{1}
    \and Shuanghao Shu
        \inst{1}
    \and Ying Li
        \inst{2}
    \and Liang Cao
        \inst{2}
    \and Huanyuan Shan
        \inst{3}
    \and Ayodeji Ibitoye
        \inst{4,5,6}
   }

   \institute{National Astronomical Observatories, Chinese Academy of Sciences,
             Beijing 100012, China; {\it nan.li@nao.cas.cn}\\
        \and
             Taiyuan University of Technology, Shanxi Province, China;\\
        \and Shanghai Astronomical Observatory, Chinese Academy of Sciences, Shanghai 200030, China;\\
        \and Department of Physics, Guangdong Technion – Israel Institute of Technology, Guangdong 515063, China;\\
        \and Centre for Space Research, North-West University, Potchefstroom 2520, South Africa;\\
        \and Department of Physics and Electronics, Adekunle Ajasin University, P. M. B. 001, Akungba-Akoko, Ondo State, Nigeria;\\
        % \and
        %      Full institute address for the third author\\
\vs\no
   {\small Received 20xx month day; accepted 20xx month day}}

\abstract{
Source extraction is crucial in analyzing data from next-generation, large-scale sky surveys in radio bands, such as the Square Kilometre Array (SKA). Several source extraction programs, including SoFiA and Aegean, have been developed to address this challenge. However, finding optimal parameter configurations when applying these programs to real observations is non-trivial. For example, the outcomes of SoFiA intensely depend on several key parameters across its preconditioning, source-finding, and reliability-filtering modules. To address this issue, we propose a framework to automatically optimize these parameters using an AI agent based on a state-of-the-art reinforcement learning (RL) algorithm, i.e., Soft Actor-Critic (SAC). The SKA Science Data Challenge 2 (SDC2) dataset is utilized to assess the feasibility and reliability of this framework. The AI agent interacts with the environment by adjusting parameters based on the feedback from the SDC2 score defined by the SDC2 Team, progressively learning to select parameter sets that yield improved performance. After sufficient training, the AI agent can automatically identify an optimal parameter configuration that outperform the benchmark set by Team SoFiA within only 100 evaluation steps and with reduced time consumption. Our approach could address similar problems requiring complex parameter tuning, beyond radio band surveys and source extraction. Yet, high-quality training sets containing representative observations and catalogs of ground truth are essential.
\keywords{methods: statistical --- software: development --- software: simulations --- radio lines: galaxies --- galaxies: statistics --- surveys}
}

   \authorrunning{X. Zhou et al.}            %author_head in even pages
   \titlerunning{AI Agent For SoFiA-2}  % title_head in odd pages

   \maketitle
%% The author head (on even pages) and the title head (on odd pages) will be
%% automatically extracted from \author{} and \title{}. Whenever the title is too long,
%% you will be asked to supply a shorter one by inserting either \authorrunning{} or
%% \titlerunning{} before \maketitle. Anyway, you can specify your own heads.
%%
%%
%% Note: In the following text body of your manuscript, please note several differences from
%%       other major journals:
%% (1) \subsection{Please Capitalize the First Letter of Each Notional Word in Subsection Title}
%% (2) Please Capitalize the First Letter of Each Notional Word in all tables' captions

%
%________________________________________________ sections below
%
\section{Introduction}           %% first-level sections will be auto-capitalized
\label{sect:intro}
Radio telescopes play a pivotal role in modern astrophysics by enabling observations of the Universe at centimeter- and meter-scale wavelengths. Unlike optical telescopes that detect photons in the visible spectrum, radio telescopes are sensitive to electromagnetic radiation in the radio frequency range, making them uniquely capable of probing neutral hydrogen gas (HI), pulsars, and large-scale structures obscured at shorter wavelengths~\citep{Pritchard2012,Liu2020,Flitter2024,Bera2023,Beskin2015,Becker2007, Brown2009, Willis1978}. Radio observations are commonly conducted using interferometers, arrays of antennas that sample the spatial coherence of incoming radio waves. The fundamental measurement in radio interferometry is the visibility, a complicated quantity that encodes spatial frequency information for a given baseline (the vector separation between two antennas), frequency channel, and time stamp. These visibilities collectively form a sparse sampling of the Fourier transform of the sky brightness distribution~\citep{Monnier2013}. Before scientific analysis can proceed, the interferometric data undergo a series of preprocessing steps to calibrate and convert into a usable form, typically an image or spectral data cube with two spatial dimensions and one spectral or frequency dimension. 

The detection of extragalactic sources in these spectral data cubes has become an essential component of modern radio astronomy. With the operation and proposition of current and next-generation interferometric surveys, such as Five-hundred-meter Aperture Spherical Telescope (FAST, ~\citet{Nan2011, Li2013}), Australian square kilometre array pathfinder (ASKAP,  ~\citet{Hotan2021,Hale2021}), MeerKAT~\citep{Jonas2016} and Square Kilometer Array (SKA, ~\citet{Dewdney2009}), the volume and complexity of data have increased dramatically. Identifying faint, spatially-extended neural hydrogen (HI) sources in these datasets requires robust, automated source finding algorithms that can operate efficiently and reliably under noise conditions and instrumental systematics. A number of source-finding algorithms have been developed and continue to be actively improved, such as Source Finding Application 2 (SoFiA-2~\footnote{\url{https://gitlab.com/SoFiA-Admin/SoFiA-2}}, ~\citet{Serra2015,Westmeier2021}), Aegean~\footnote{\url{https://github.com/PaulHancock/Aegean}}~\citep{Hancock2012, Hancock2018}, LiSA~\footnote{\url{https://github.com/epfl-radio-astro/LiSA}}~\citep{Tolley2022}.

SoFiA-2 is a widely used software developed to detect and characterize HI sources in three-dimensional spectral data cubes. It includes a modular pipeline that supports a range of preprocessing techniques, such as noise normalization, background subtraction, a robust source detection algorithm, and post-processing tools such as reliability filtering. However, the performance of SoFiA-2 is highly dependent on a number of user-defined parameters, including smoothing kernels, detection thresholds, reliability cuts, which influence both the completeness and purity of the final catalogue. Optimal tuning of these parameters is crucial for maximizing scientific return but is often performed manually or through computationally expensive grid searches. 

Fortunately, Artificial intelligence (AI) agents can address this challenge. AI agents are designed to interact with and learn from their environments by making sequential decisions that maximize long-term goals~\citep{Adam2025,Durante2024}. A central framework for developing such agents is reinforcement learning (RL, ~\citet{kaelbling1996, sutton1992, Ghasemi2024, Jaeger2023}), where an agent learns to optimize its actions based on reward feedback through interactions with an environment modeled as a Markov Decision Process (MDP, ~\citet{white1989}). In this paradigm, the agent gradually improves its policy by exploring possible strategies and exploiting knowledge gained from past experiences. Over the past decade, several key RL algorithms have been developed to address different types of learning tasks. Deep Q-Networks (DQN, \citet{Mnih2013,roderick2017,fan2020}), for instance, extend classical Q-learning by using deep neural networks to approximate action-value functions, achieving impressive results in discrete action spaces such as Atari games. However, DQN and its variants are limited to discrete action settings and suffer from instability in function approximation. To tackle problems in continuous action spaces, algorithms such as Deep Deterministic Policy Gradient (DDPG, ~\citet{Lillicrap2015}) and Proximal Policy Optimization (PPO, ~\citet{Schulman2017}) have been proposed. DDPG combines deterministic policy gradients with actor-critic architecture and experience replay, enabling it to scale to high-dimensional control tasks. Despite its success, DDPG is known to be sensitive to hyperparameters and exploration noise, often requiring extensive tuning~\citep{Matheron2019,Tiong2020}. PPO, on the other hand, introduces a clipped surrogate objective to ensure stable updates to the policy, making it more robust and widely adopted, especially in environments with complex dynamics. However, as an on-policy algorithm, PPO suffers from lower sample efficiency compared to off-policy approaches~\citep{Yuan2025}. To address these limitations, the Soft Actor-Critic (SAC, ~\citet{Haarnoja2018}) algorithm has been introduced as a state-of-the-art method for continuous control. SAC builds on the actor-critic framework but incorporates a key innovation: the maximum entropy reinforcement learning principle. By augmenting the standard reward objective with an entropy term, SAC explicitly encourages stochastic policies that promote exploration. This results in agents that are more robust to uncertainty and less prone to premature convergence. SAC is also off-policy, allowing for more efficient reuse of past experiences. Furthermore, SAC’s combination of multi-step Q-value targets, target networks, and soft updates contribute to improved convergence and resilience in training. 

The reinforcement learning algorithms have gained significant attention in the field of astronomy, particularly in the areas of telescope engineering, such as adaptive optical system control~\citep{Landman2020,Landman2021,Nousiainen2021,Nousiainen2022,Nousiainen2024,Parvizi2023, Pou2024}, autonomous observation scheduling~\citep{Terranova2023}, and instrument wavefront correction~\citep{Gutierrez2024deep, Gutierrez2024image}. Beyond these applications, RL has also demonstrated its potential in scientific research, such as connecting galaxy properties and dark matter halos~\citep{Moster2021}, reconciling Hubble tension~\citep{Sharma2024}, and addressing non-LTE problems in solar physics~\citep{Panos2025}. Furthermore, RL can be effectively employed in data preprocessing pipelines for radio and optical observations~\citep{Yatawatta2021,Yatawatta2023,Kirk2025}. For further insights into applications of reinforcement learning in astronomy, please refer to~\citet{Yatawatta2024reinforcement}.

In our work, following the effort of optimizing parameters by RL for photometry measurement using SExtractor~\citep{Bertin1996} in optical observations~\citep{Jia2025}, we propose an approach to parameter optimization in SoFiA-2. Given the increased number of parameters, we instead adopt SAC algorithm, which enables to search multiple parameters and in continuous spaces. In our framework, the agent interacts with a simulation-based environment in which it selects SoFiA-2 parameter configurations, evaluate the resulting score against a known ground truth, and receives rewards based on the score. Through repeated interaction and policy updates, the SAC agent learns to navigate the high-dimensional parameter space and discover configurations that optimize source detection performance. We validate our approach using synthetic HI data cubes from SKA Science Data Challenge 2 (SKA-SDC2~\footnote{\url{https://sdc2.skao.int/}}, ~\citet{Hartley2023}), where ground-truth source catalogs are available. Our results demonstrate that SAC-based tuning strategy can yield source catalogs with higher scores compared to manual tuning and random search, and can be generalized to unseen data. This study demonstrates the applicability of AI agents for optimal parameter selection in scientific software and highlights their potential in the era of big-data astronomy.

This paper is organized as follows. In Section~\ref{sec: data}, we introduce the datasets used in this study, including the HI data cube and the corresponding truth catalog from SDC2. Section~\ref{sec: method} provides an overview of SoFiA-2 and the reinforcement learning framework, SAC algorithm. The training procedures are detailed in Section~\ref{sec: training}, while Section~\ref{sec: results} presents the results obtained on both the development and large development datasets. In Section~\ref{sec: discussion}, we discuss parameter importance, the scenario where the agent is trained on different configurations of data, potential applications in real observations, limitations, and offer practical tips for training and deploying the AI agent. Finally, we conclude the paper in Section~\ref{sec: conclusion}. Data availability is summarized in Section~\ref{sec: data avail}. Appendix~\ref{app: sky} provides details on the detected source catalog and compares source properties with those in the truth catalog.

%% Authors can give a citation as 'Michel et al. 1992'.
%% You may also use \cite, \citep and \citet for citation, and use Table~1 or Figure~1
%% and so forth. Using \ref and \label for cross-references of Tables/Figures
%% is a good way in adjusting/adding/removing text, tables or figures.

\section{Data}\label{sec: data}
SKA-SDC~\footnote{\url{https://www.skao.int/en/science-users/160/data-challenges}} is a regular science preparatory activities for the science community. These challenges aim to inform the development of the data reduction and analysis workflows, allowing the community to become familiar with the standard data that SKA will produce. The second data challenge involves source finding and characterization from a simulated spectral line observation that represents the SKAO view of HI emission up to $z=0.5$~\citep{Hartley2023}. The whole simulated HI data product covers approximately 20 square degrees of sky. The pixel scale is $2.8\times2.8\ \rm arcsec/pix$, and beam size is $7\ \rm arcsec$.  The bandwidth is from 950-1150 MHz, with a resolution of 30 kHz, corresponding to redshift interval $z=0.235-0.495$. The 3D data cube has $5853\times5851\times6668$ pixels in two spatial and wavelength directions. To mimic the real observations, noise level of a 2,000 hour observation and systematics encompassing imperfect continuum subtraction, radio frequency interference (RFI) flagging and excess noise due to RFI are also included. 

In addition to the full dataset, two smaller datasets, each with a coverage area of 1 and 0.25 square degrees, are generated through the same methodology employed for the full dataset. These two datasets are associated with truth catalogs, facilitate the development of source finding algorithms. For the construction of our AI agent, we employ the smaller dataset as the training and validation dataset, while the larger one is employed for testing purpose. This decision is intended to restrict the execution time of SoFiA-2, as it will be executed repeatedly during the training of our agent.

\section{Methodology}\label{sec: method}
This section outlines the configuration of SoFiA-2 and presents the reinforcement learning algorithm, SAC, employed in constructing the AI agent.
\subsection{SoFiA-2}\label{sec: sofia}
SoFiA-2~\citep{Serra2015, Westmeier2021} is a modular and scalable software package designed for the automated detection and analysis of sources in three-dimensional spectral data cubes, with a specific focus on extragalactic HI surveys. As a complete redesign and re-implementation of the original pipeline in C language, SoFiA-2 addresses the increasing computational requirements of current and forthcoming large-scale HI surveys. 

The workflows of SoFiA-2 encompass several procedures, including preconditioning, source finding, linker, and reliability measurement. These modules are controlled by several parameters defined in a parameter configuration file. To reduce computational costs, SoFiA-2 can accept a portion of the data cube by specifying~\texttt{input.region} in pixel coordinates. Additional auxiliary data, such as masks and weights, can also be inputted. The preconditioning process involves several preprocessing steps prior to source finding, including data flagging, continuum subtraction, noise normalization, and background subtraction~\citep{Monnier2013,Hanssen2001}. Data flagging is employed to eliminate interference and artifacts. Various flagging methods can be utilized, such as region-based, catalog-based, and threshold-based approaches. Continuum subtraction aids in removing low-level residual continuum emission using a robust polynomial fitting algorithm. If noise levels vary across the data cube, the noise normalization module measures and corrects the noise in both spectral and spatial directions. Background subtraction, also known as ripple filter, is used to remove spatially and spectrally extended artifacts that are variable across the sky or frequency. Each preconditioning process can be independently enabled and disabled, providing flexible preprocessing procedures. 

The source finding algorithm employed by SoFiA-2 is the S + C finder. This algorithm operates by smoothing the input data cube on multiple scales in both spatial and spectral directions, as specified by~\texttt{scfind.kernelsXY} and~\texttt{scfind.kernelsZ}. SoFiA-2 applies Gaussian filters and boxcar filters in the spatial and spectral domains, respectively. It is important to note that the width of the boxcar filter must be odd in units of channels. The filters from both domains are combined to generate $N\times M$ distinct combinations, where $N$ and $M$ represent the number of spatial and spectral filters, respectively. These combinations are then successively applied to the data cube, and all pixels with a flux density exceeding~\texttt{scfind.threshold} times the RMS noise level are subsequently masked. 

Following the source finding, the linker operates by aggregating the mask pixels into distinct sources. This algorithm possesses several parameters, including \texttt{linker.radiusXY} and \texttt{linker.radiusZ}, which determine the merging length in spatial and spectral directions, respectively. To mitigate the occurrence of false detections caused by noise peaks and artifacts, a minimum requirement of pixel sizes for sources should also be specified by \texttt{linker.minSizeXY} and \texttt{linker.minSizeZ}. 

The subsequent step involves the calculation of reliability~\citep{Serra2012}, which provides an automated approach to determine the reliability of detections and discard sources that fail to meet the reliability threshold specified by \texttt{reliability.threshold}. In addition to the threshold, the \texttt{reliability.scaleKernel} defines the scale factor for the Gaussian kernel employed by the reliability algorithm to measure the density of positive and negative detections within parameter space. Furthermore, \texttt{reliability.minSNR} establishes an additional threshold to further eliminate false detections that cannot be filtered by the reliability calculation. For more details in the meaning of parameters and settings, please refer to the user manual~\footnote{\url{https://gitlab.com/SoFiA-Admin/SoFiA-2/-/wikis/documents/SoFiA-2_User_Manual.pdf}} of SoFiA-2. 

Following the parameter settings established by Team SoFiA~\footnote{\url{https://gitlab.com/SoFiA-Admin/SKA-SDC2-SoFiA}}~\citep{Hartley2023} in SDC2, we enable noise normalization in each spectral channel and search for \texttt{scfind.threshold}, \texttt{reliability.threshold}, \texttt{reliability.scaleKernel}, and \texttt{reliability.minSNR}. All other parameters are configured identically to their work, including \texttt{scfind.kernelsXY}, \texttt{scfind.kernelsZ}, \texttt{linker.radiusXY}, \texttt{linker.radiusZ}, \texttt{linker.minSizeXY}, and \texttt{linker.minSizeZ}. It is noted that the flagging of bright continuum sources in their work is not utilized, as its impact is minimal. The parameters taken by Team SoFiA and the considered range for each parameter in our work are presented in Table~\ref{tab: params}. Furthermore, to facilitate the measurement of source properties, \texttt{parameter.physical} option is also enabled.

\begin{table}
\centering
\caption{Considered parameters for SoFiA-2 from Team SoFiA and our work.}
\label{tab: params}
\begin{tabular}{|l|cc|}
\hline
Parameters              & \multicolumn{1}{c|}{Team SoFiA} & Our work     \\ \hline\hline
scfind.threshold        & \multicolumn{1}{c|}{3.8}        & 3.5$\sim$4.0 \\ \hline
scfind.kernelsXY        & \multicolumn{2}{c|}{0, 3, 6}                   \\ \hline
scfind.kernelsZ         & \multicolumn{2}{c|}{0, 3, 7, 15, 31}           \\ \hline
linker.radiusXY         & \multicolumn{2}{c|}{2}                         \\ \hline
linker.radiusZ          & \multicolumn{2}{c|}{2}                         \\ \hline
linker.minSizeXY        & \multicolumn{2}{c|}{3}                         \\ \hline
linker.minSizeZ         & \multicolumn{2}{c|}{3}                         \\ \hline
reliability.threshold   & \multicolumn{1}{c|}{0.1}        & 0.0$\sim$0.6 \\ \hline
reliability.scaleKernel & \multicolumn{1}{c|}{0.3}        & 0.1$\sim$0.7 \\ \hline
reliability.minSNR      & \multicolumn{1}{c|}{1.5}        & 0.0$\sim$2.0 \\ \hline
\end{tabular}
\end{table}

\subsection{AI Agent}\label{sec: agent}
Our AI agent is built upon a state-of-the-art RL algorithm, SAC~\citep{Haarnoja2018}. This algorithm builds upon the maximum entropy reinforcement learning framework, which modifies the standard reinforcement learning objective by incorporating an entropy term. The fundamental principle is to maximize both expected return and policy entropy simultaneously. In standard reinforcement learning, the objective is to find a policy $\pi(a|s)$ that maximizes the expected sum of rewards:
\begin{equation}
    J_{\rm standard}(\pi) = \mathbb{E}_{\tau\sim\pi}\left[\sum_0^{\infty}\gamma^t r(s_t, a_t)\right]
\end{equation}
where $\gamma\in[0, 1)$ is the discount factor, which determines how much the agent values future rewards compared to intermediate rewards. A smaller $\gamma$ makes the agent more short-sighted, focusing on intermediate rewards, while a value closer to 1 makes the agent more far sighted, emphasizing long-term gains. $r(s_t, a_t)$ indicates the reward function, providing a scalar feedback signal for taking action $a_t$ in state $s_t$. SAC maximizes the expected entropy-augmented return:
\begin{equation}
    J_{\rm SAC}(\pi) = \mathbb{E}_{\tau\sim\pi}\left[\sum_0^{\infty}\gamma^t r(s_t, a_t) + \alpha\mathcal{H}(\pi(\cdot|s_t))\right]
\end{equation}
where $\mathcal{H}(\pi(\cdot|s_t)) = \mathbb{E}_{a_t\sim\pi}[-\log\pi(a_t|s_t)]$ is the entropy of the policy at state $s_t$, and $\alpha > 0$ is the temperature parameter that determines the trade-off between exploration (via entropy maximization) and exploitation (via reward maximization). This formula encourages the agent to explore by favoring more stochastic policies, especially in regions where the value function is uncertain. {Since direct backpropagation through a stochastic sampling process is impossible, the stochastic policy $\pi_{\phi}(a|s)$ is typically modeled using a Gaussian distribution with a reparameterization trick, i.e., $a=\tanh(\mu_{\phi}(s) + \sigma_{\phi}(s)\cdot\epsilon)$, where $\mu$ and $\sigma$ are mean and scale for Gaussian, and $\epsilon$ represent standard Gaussian distribution, $\epsilon\sim\mathcal{N}(0, I)$. By reformulating the action in this way, the stochasticity is isolated in the $\epsilon$ term, enabling the use of standard backpropagation to update the policy network based on the expected return. Additionally, the $\tanh$ transformation is to bound the actions within a specific range, typically [-1, 1], and to stabilize the maximum entropy objective. }

SAC employs two soft Q-value estimators, $\mathcal{Q}_{\theta_1}(s, a)$ and $\mathcal{Q}_{\theta_2}(s, a)$, to mitigate overestimation bias in value estimates, a known issue in function approximation for Q-learning. These two estimators can also be called as critics. During training, target Q-values are computed using the minimum of these two critics to form a conservative estimate:
\begin{equation}
    \mathcal{Q}_{\rm target}(s, a) = \min(\mathcal{Q}_{\theta_1}(s, a), \mathcal{Q}_{\theta_2}(s, a))
\end{equation}
To stabilize training, target networks $\mathcal{Q}_{\theta^\prime_1}(s, a)$ and $\mathcal{Q}_{\theta^\prime_2}(s, a)$ are maintained as exponentially moving averages of the main Q-networks using Polyak averaging:
\begin{equation}
    \theta^\prime_i \leftarrow\tau\theta_i+(1 - \tau)\theta_i^\prime,\quad i\in\{1, 2\}
\end{equation}
where $\tau \ll 1$ controls the update rate. Each Q-estimator is updated by minimizing the soft Bellman residual:
\begin{equation}
    \mathcal{L}_{\mathcal{Q}}(\theta_i) = \mathbb{E}_{(s, a, r, s^\prime)\sim\mathcal{D}}\left[(\mathcal{Q}_{\theta_i}(s, a) - (r + \gamma\mathbb{E}_{a^\prime\sim\pi}[\mathcal{Q}_{\rm target}(s^\prime, a^\prime) - \alpha\log\pi(a^\prime|s^\prime)]))^2\right]
\end{equation}
where $\mathcal{D}$ is the replay buffer for experiences. Additionally, an actor network is introduced to output a stochastic policy, modeled as a Gaussian distribution from which actions are sampled. This network is updated to minimize the KL divergence between itself and the Boltzmann distribution induced by the Q-estimator:
\begin{equation}
    \mathcal{L}_{\pi}(\phi)=\mathbb{E}_{s\sim\mathcal{D}}\left[\mathbb{E}_{a\sim\pi_{\phi}(\cdot|s)}[\alpha\log\pi_{\phi}(a | s) - \mathcal{Q}_\theta(s, a)]\right]
\end{equation}
This encourages the policy to select actions that both maximize expected return and maintain high entropy, thereby ensuring sufficient exploration. The temperature parameter $\alpha$ can be fixed or automatically tuned. In the latter case, $\alpha$ is optimized to match a target entropy $\mathcal{H}_{\rm target}$, using the objective:
\begin{equation}
    \mathcal{L}(\alpha)=\mathbb{E}_{\alpha\sim\pi}[-\alpha(\log\pi(a|s) + \mathcal{H}_{\rm target}])]
\end{equation}
which allows the agent to dynamically balance exploration and exploitation during training. 

In our work, we implement the agent using \texttt{gymnasium}~\footnote{\url{https://gymnasium.farama.org/}} and \texttt{stable\_baselines3}~\footnote{\url{https://stable-baselines3.readthedocs.io/en/master/}}. The former is utilized to establish a reinforcement learning environment for optimal parameter searching of SoFiA-2, while the latter offers a predefined SAC algorithm. The policy model is configured to be “MlpPolicy,” indicating that the networks employed in SAC are all multi-layer perceptrons (MLP), which aligns well with our objectives. 

% \usepackage{multirow}
% \usepackage{multirow}
\begin{table}
\centering
\caption{The pixel range, number of sources, SDC2 score and number of matched sources by Team SoFiA for the training and testing data.}
\label{tab: data sets}
\begin{tabular}{|c|c|c|c|c|}
\hline
                               & Range             & $N_{\rm sources}$ & $S_{\rm Team\ SoFiA}$ & $N_{\rm matched,\ Team\ SoFiA}$ \\ \hline\hline
\multirow{3}{*}{Training data} & 0, 322, 0, 322     & 572               & 38.69                & 61                            \\ \cline{2-5} 
                               & 322, 643, 0, 322   & 660               & 36.42                & 52                            \\ \cline{2-5} 
                               & 322, 643, 322, 643 & 734               & 60.2                 & 78                            \\ \hline
Testing data                   & 0, 322, 322, 643   & 717               & 44.37                & 62                            \\ \hline
\end{tabular}
\end{table}

\section{Training}\label{sec: training}
To improve training efficiency and ensure data representativeness, we selectively utilize three patches from the development dataset as training data, each covering an area of $0.0625$ square degrees. These patches are selected to cover both sparse and dense fields represented by number of sources as outlined in Table~\ref{tab: data sets}. Further investigations on the training data are discussed in Section~\ref{sec: training sets}. The rest patch is considered to be the testing data. During training, all three patches are processed by SoFiA-2 simultaneously, using a shared set of parameters. This strategy is equivalent to conducting training in a larger complete sky patch. The considered parameters mentioned in Section~\ref{sec: sofia} are randomly initialized at the beginning of training, while in the subsequent steps, they are determined by actions output by the MLP in SAC algorithm.

The performance of source detection is measured using the SDC2 Scorer~\footnote{\url{https://gitlab.com/ska-telescope/sdc/ska-sdc/-/tree/master/ska_sdc}}~\citep{Hartley2023}, which takes both the detected source catalog and the truth catalog as input. This score evaluates the performance based on two factors: the number of matched sources (in terms of positions in 3D cube, identified by RA, Dec, and central frequency, $\nu$) and the consistency of physical properties between matched sources, including HI size, $S$, integrated line flux $F$, position angle $\theta$, inclination angle, $i$, and line width, $w_{\rm 20}$. These quantities are derived from output catalog of SoFiA-2 using the conversion method provided by Team SoFiA~\footnote{\url{https://gitlab.com/SoFiA-Admin/SKA-SDC2-SoFiA/-/blob/master/scripts/physical_parameter_conversion.py}}. This score facilitates a comprehensive and reliable assessment of source detection quality.  The highest score is achieved when all sources are matched in both position and properties within predefined thresholds, yielding a score equal to the total number of sources in the truth catalog. For further detailed information on the scoring algorithm, we refer interested reader to~\citet{Hartley2023}. To account for variations in source counts across different patches, we define a score ratio, SR, by normalizing score to the the total number of sources. For the three training patches, the overall score ratio is calculated as the sum of individual score in each patch divided by the total number of ground-truth sources. Details of the training and testing patches, including pixel ranges, number of sources, benchmark scores and number of matched sources obtained through parameter set of Team SoFiA is outlined in Table~\ref{tab: data sets}.

The SAC algorithm operates as an off-policy, maximum entropy, actor-critic framework that learns from a replay buffer filled with its own past experiences, as discussed in Section~\ref{sec: agent}. Each experience consists of a tuple containing the state, action, done flag, next state, and reward, all collected through interactions with the environment during training. In our case, the environment corresponds to the SoFiA-2 and the simulated HI cubes. The state is constructed using the current values of the selected parameters, along with the corresponding SR defined above. The action, generated by the Actor network, determines the update amplitude for each parameter. The done flag signals the termination of an episode, after which a new episode is initialized. In our implementation, the done flag is triggered only when the maximum number of steps for an episode is reached, allowing for comprehensive exploration of the parameter space. The reward, on the other hand, functions as the guiding signal for learning, encouraging the agent to transition toward parameter configurations that yield improved performance. The reward is computed based on SR and is defined as follows:
\begin{equation}
    R = 
    \begin{cases}
        -5.0 &\ \text{No output from SoFiA or SR \textless\ 0,}\\
        10\times e^{5.0\times {\rm SR} - 1.0} &\ \text{base reward,}\\
        +2.0 &\ \text{positive improvement for SR,}\\
        -2.0 &\ \text{negative improvement for SR,}\\
        +3.0 &\ \text{bonus for reaching highest SR,}\\
        -3.0 &\ \text{penalty for exhausting max steps.}
    \end{cases}
\end{equation}
The first condition penalizes parameter sets that lead to no detected sources or severely degraded performance. The second term provides a scaled base reward that increases exponentially with the SR, encouraging better overall performance. The third and fourth conditions offer additional bonus or penalties based on relative improvement or degradation between steps. An extra bonus is granted when the highest SR is achieved, while a penalty is applied for exhausting the maxmimum number of steps in an episode, encouraging the agent to find optimal parameters efficiently. This reward formulation provides a clear and structured path for guiding the optimization of SoFiA-2 parameters. Using these rewards, the Critic networks learn the Q-function, which estimates the expected cumulative reward for taking a given action in specific state. In SAC, two Q-networks are used to stabilize learning and reduce overestimation bias. 

The total number of training steps is set to 10,000, with each episode having a maximum of 100 steps. The replay buffer used to store experiences has a capacity of 1,000 entries. During training, random samples from this buffer are drawn to update the networks. As training progresses, older experiences are gradually replaced by new ones, enabling the agent to converge toward an optimal parameter set for SoFiA-2. The agent is trained using a batch size of 64, meaning that training begins once 64 experiences have been collected. All other key hyperparameters, including the learning rate, soft update coefficient ($\tau$), discount factor ($\gamma$), model training frequency, and number of gradient steps, are configured using the default settings provided by the SAC implementation in the \texttt{stable\_baselines3} library.

\section{Results}\label{sec: results}
In this section, we demonstrate and analyze the results of our agent on development and large development datasets mentioned in Section~\ref{sec: data}.

\subsection{Development dataset}\label{sec: development}

% \begin{figure*}
%     \centering
%     \includegraphics[width=\linewidth]{total_score_3regions.png}
%     \caption{Total scores for three training regions with respect to steps in training. The benchmark by Team SoFiA is displayed in red dashed line.}
%     \label{fig: score to steps}
% \end{figure*}

\begin{figure*}
    \centering
    \includegraphics[width=0.49\linewidth]{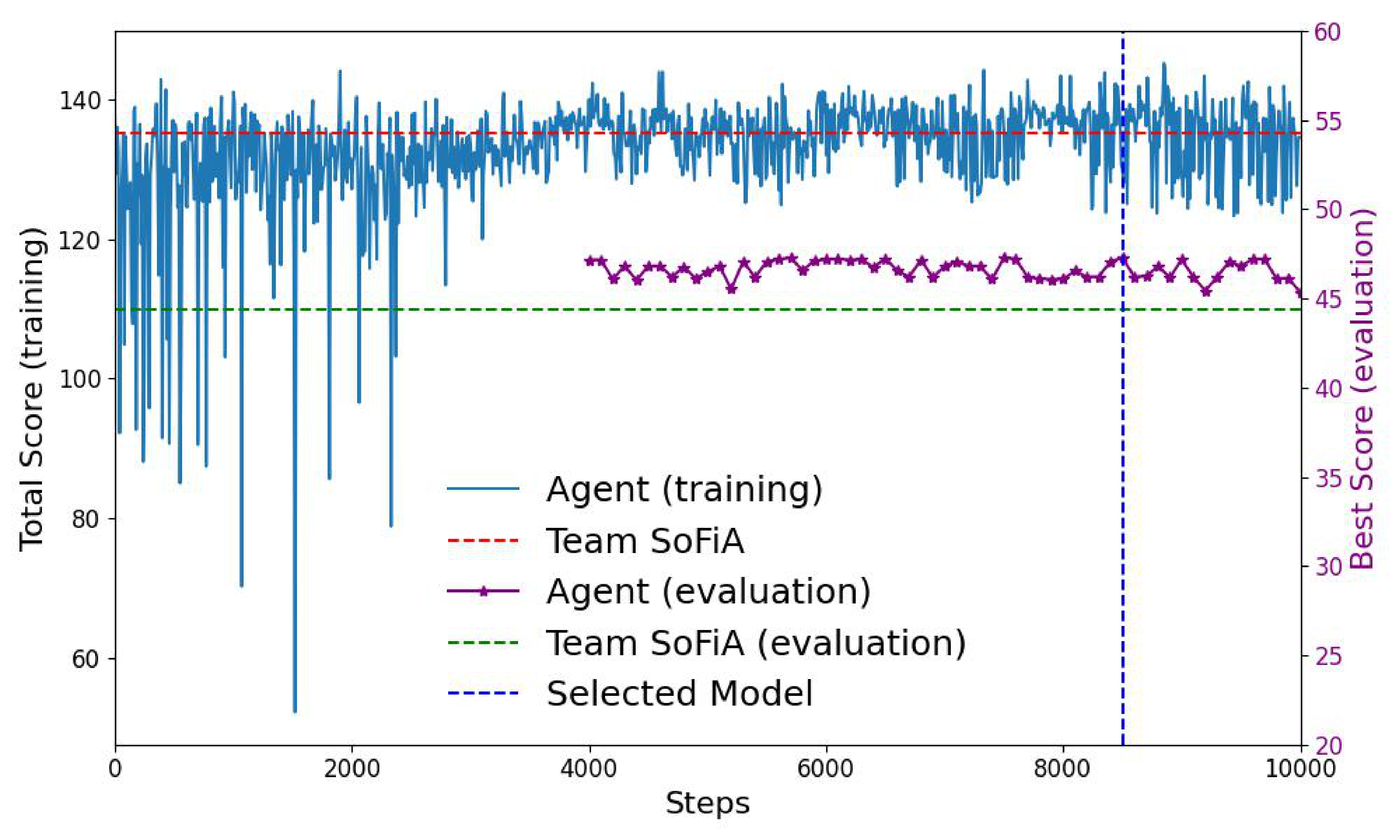}
    \includegraphics[width=0.49\linewidth]{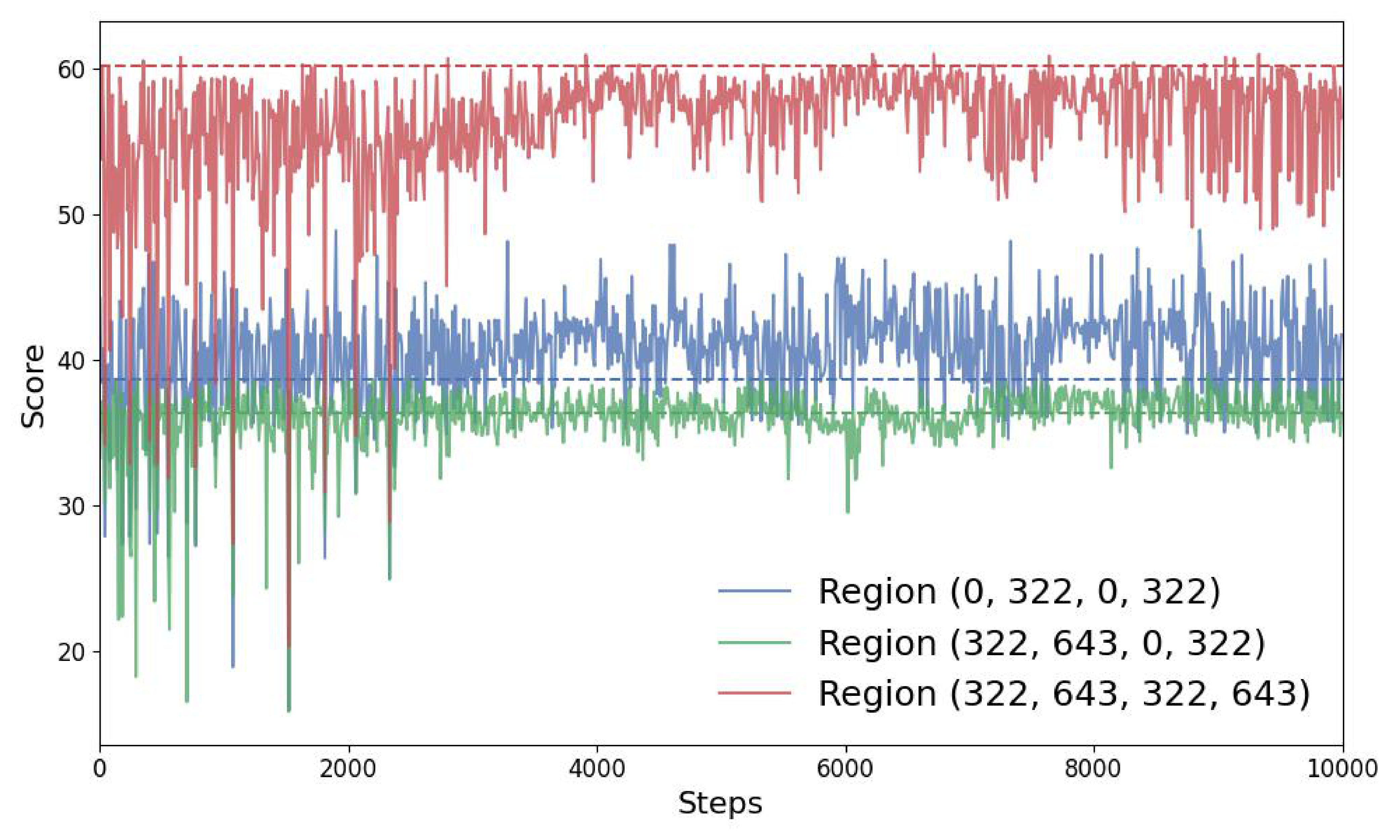}
    \caption{\textit{Left}: Total scores for the three training patches as a function of training steps. The benchmark score achieved by Team SoFiA is shown as a red dashed line. The highest evaluation scores obtained from the testing patch, using models saved every 100 training steps, are shown in purple. The benchmark for the testing patch is indicated by the green dashed line. The blue dashed line marks the best-performing model, selected at step 8,500. \textit{Right}: Training scores over steps for each of the three individual patches. The benchmark score for each patch is indicated by the corresponding dashed line in same color.}
    \label{fig: score regions}
\end{figure*}

\begin{figure*}
    \centering
    \includegraphics[width=\linewidth]{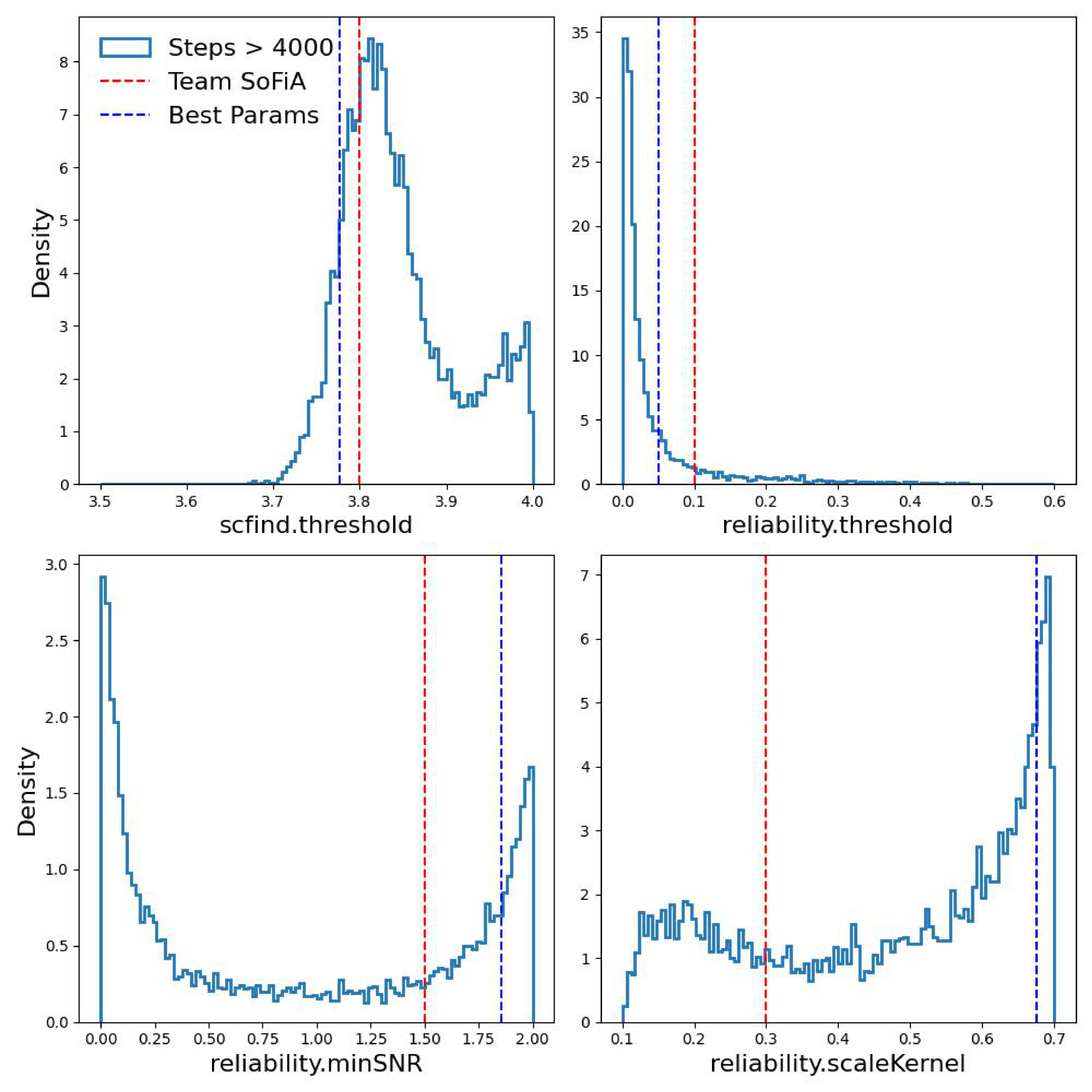}
    \caption{The distributions for the four parameters after 4,000-step training. The parameters from Team SoFiA and the best parameters from agent are also displayed in red and blue dashed lines respectively.}
    \label{fig: params dist}
\end{figure*}

\begin{figure*}
    \centering
    \includegraphics[width=\textwidth]{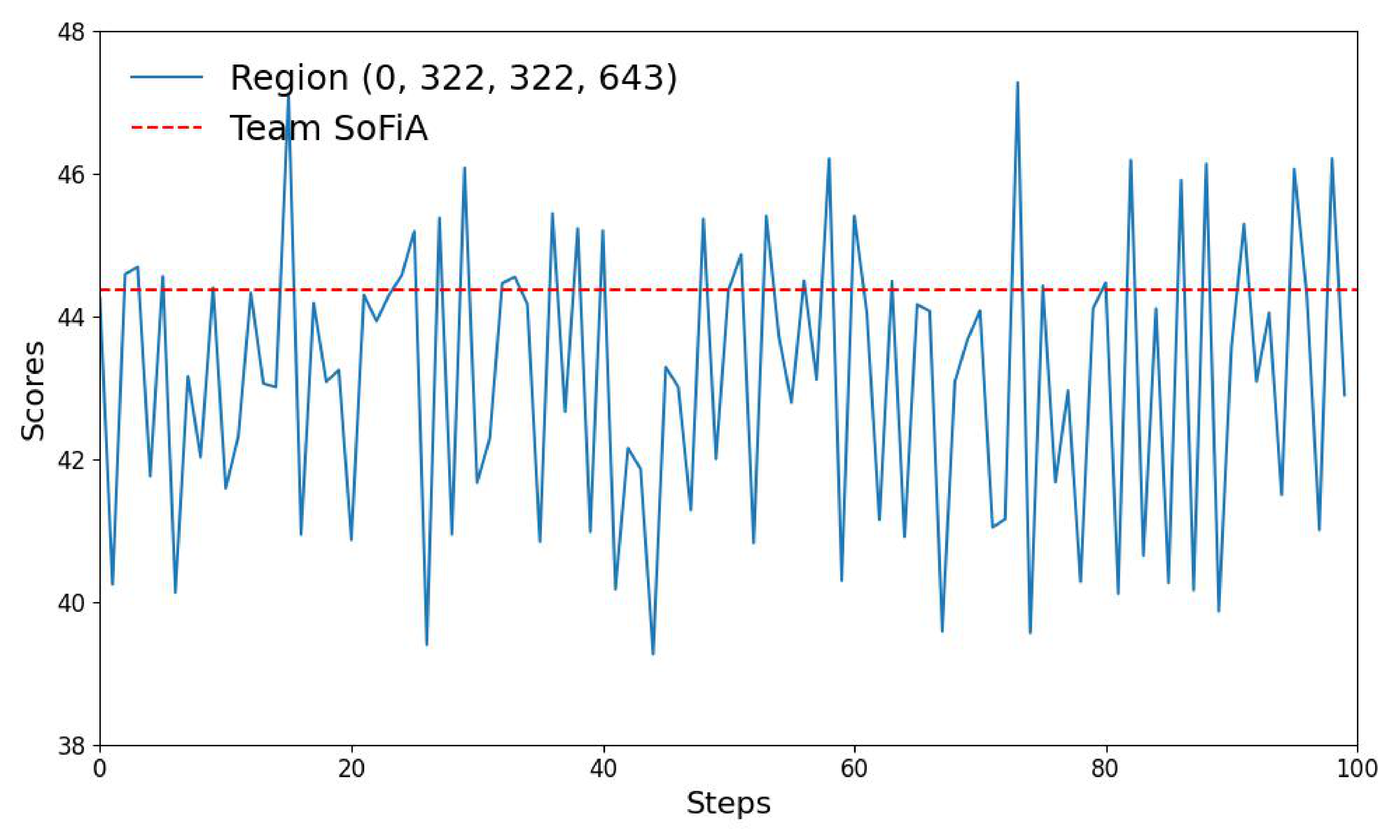}
    \caption{Scores for testing patch with respect to steps through the best model at step 8,500. The benchmark score is displayed in red dashed line. }
    \label{fig: scores eval model}
\end{figure*}

The training of agent is performed on three patches of development dataset as mentioned in Section~\ref{sec: training}. The total scores in these patches as a function of training steps are shown in the left panel of Figure~\ref{fig: score regions}. For clarity, scores are sampled every 10 steps, and the red dashed line represents the benchmark score achieved by parameters from Team SoFiA. We observe that the score stabilizes after approximately 4,000 steps, fluctuating around the benchmark level. The right panel of Figure~\ref{fig: score regions} further illustrates the scores in the three training patches individually as discussed in Section~\ref{sec: training}. And the benchmark score for the corresponding region is indicated by a dashed line of the same color. For patch (0, 322, 0, 322), shown in blue, the agent consistently surpasses the benchmark. In patch (322, 643, 0, 322), shown in green, the score closely tracks the benchmark. However, in patch (322, 644, 322, 643), shown in red, the score rarely exceeds the benchmark, suggesting that performance beyond this level is particularly challenging in this field. To gain insight into the learned parameter preferences, we examine the distribution of parameter values used in steps beyond 4,000 and present them in Figure~\ref{fig: params dist}. The results reveal that the agent's prefered values for  \texttt{reliability.threshold}, \texttt{reliability.minSNR}, and \texttt{reliability.scaleKernel} notably deviate from those selected by Team SoFiA. In contrast, the agent’s choice for \texttt{scfind.threshold} closely aligns with the benchmark configuration.

We evaluate the generalizability of the trained agent using a testing patch from the development dataset, as listed in Table~\ref{tab: data sets}. The evaluation is conducted using SAC models saved at every 100 training steps after step 4,000. For each model, the agent is run for up to 100 steps to obtain the best parameter sets, with highest score achieved. These scores are recorded and shown in purple in the left panel of Figure~\ref{fig: score regions}. And for comparison, the benchmark score of this patch is also displayed. We notice that the best scores consistently exceed the benchmark, indicating that the agent has generalization capability. Among the evaluated models, the one at step 8,500 achieves the highest score and is selected for further analysis. Figure~\ref{fig: scores eval model} shows the score progression for this model during evaluation. As expected, the agent is able to achieve the optimal score within 100 steps, demonstrating its ability to generalize effectively to previously unseen data.

% \begin{figure*}
%     \centering
%     \includegraphics[width=\linewidth]{scores_in_ldev_regions_322.png}
%     \caption{The scores with respect to steps for agent on each patch with (322, 322, 6668) of large development datasets. The red dashed lines represent the benchmark score by Team SoFiA.}
%     \label{fig: agent ldev 322}
% \end{figure*}

\begin{figure*}
    \centering
    \includegraphics[width=\textwidth]{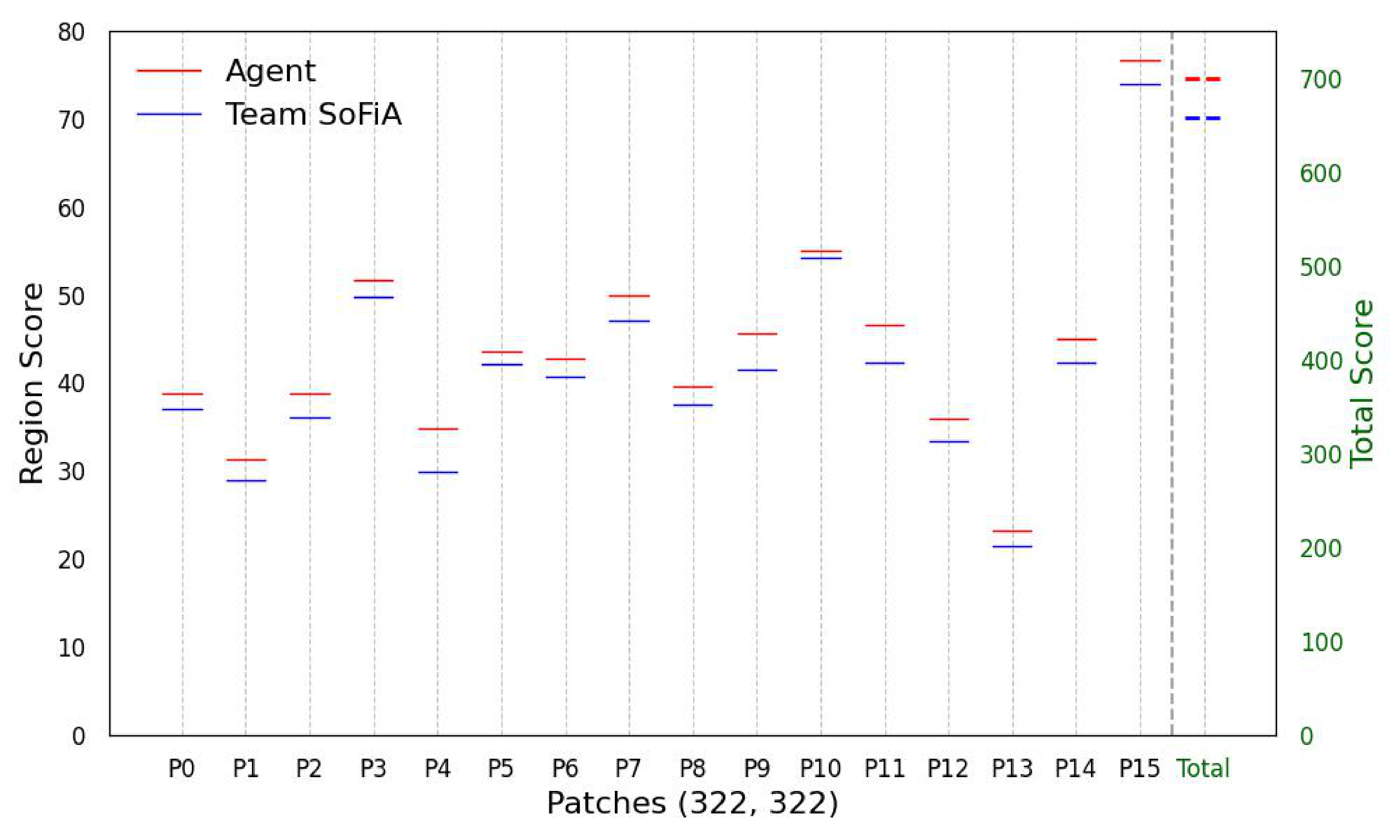}
    \caption{The scores in each patch by agent (red) and benchmark (blue) for large development dataset. The aggregated score is also shown by the right y-axis. {The x-axis denotes the patch IDs with size (322, 322) in pixels.}}
    \label{fig: ldev 322}
\end{figure*}

\subsection{Large development dataset}\label{sec: development large}
The large development dataset spans approximately 1 square degree, with a data array size of (1286, 1286, 6668). Following the same approach as with the testing dataset, we apply the trained agent to each patch of area $0.0625$ square degrees, corresponding to a patch in spatial size of (322, 322). For each patch, the agent is allowed up to 100 evaluation steps to identify the best-performing parameter set. The best score achieved by the agent for each patch, along with the corresponding benchmark score from Team SoFiA, is shown in Figure~\ref{fig: ldev 322}, where red and blue lines represent the agent and benchmark scores respectively. {The x-axis denotes the patch IDs with size (322, 322) in pixels. }
% The x-axis denotes the pixel range, which can be restored as $(i\times S_{\rm patch}, (i + 1)\times S_{\rm patch}, j\times S_{\rm patch}, (j + 1)\times S_{\rm patch})$. $i$ and $j$ are the tick labels in x-axis and $S_{\rm patch}$ indicates the patch size, as 322 here. 
While the performance improvements vary across patches, the agent consistently identifies parameter sets that outperform the benchmark in every patch. Additionally, the cumulative score across all patches in the large development dataset is shown on the right y-axis of the figure. These results demonstrate that the agent is capable of generalizing to unseen sky areas and effectively optimizing parameters for improved source-finding performance with SoFiA-2. 

\section{Discussion}\label{sec: discussion}
In this section, we discuss the importance of individual parameters, examine the scenario where the agent is trained on different configurations of data, and explore potential applications to real observational data. Finally, we emphasize the limitations of our agent. 
\begin{figure*}
    \centering
    \includegraphics[width=\textwidth]{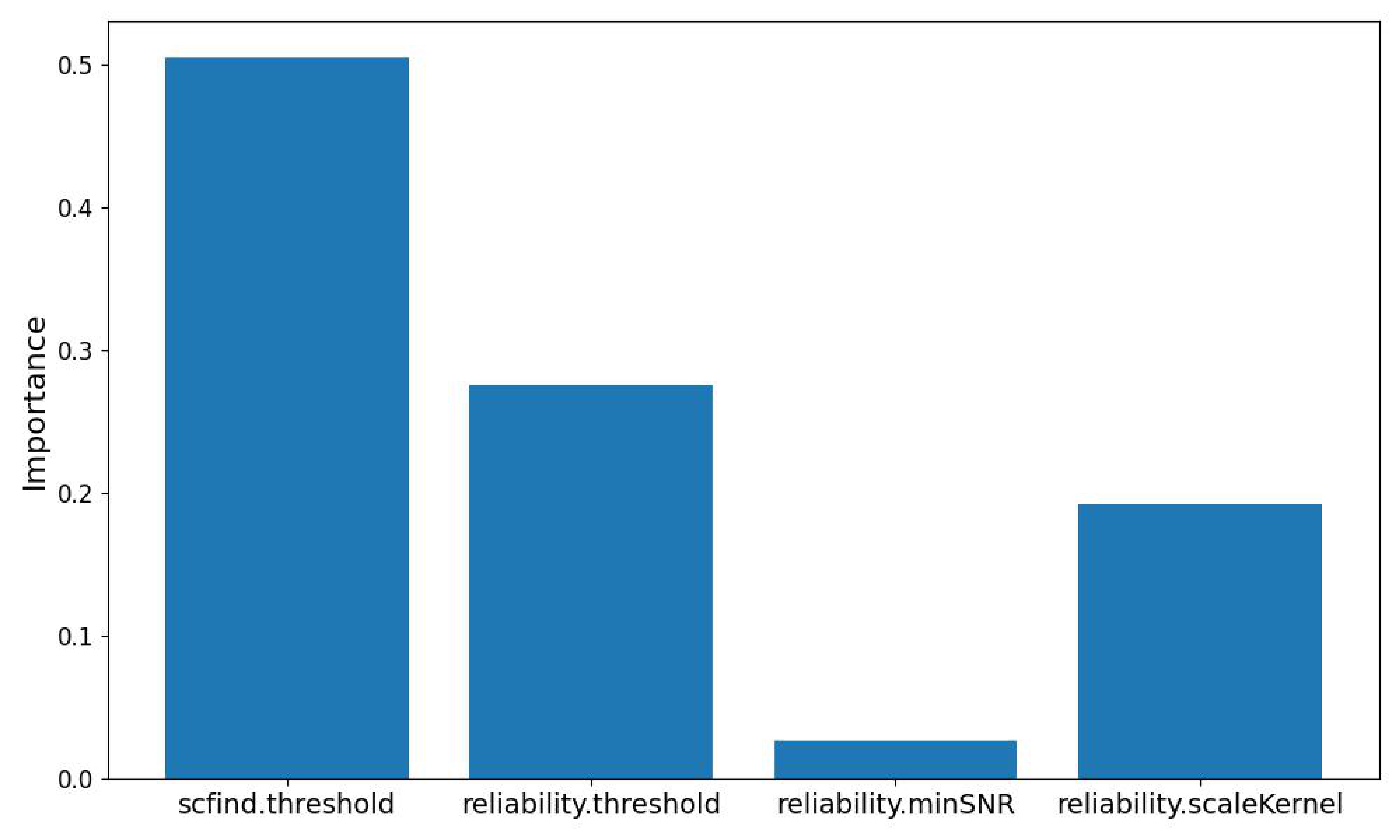}
    \caption{The importance of each parameter by RF regressor. }
    \label{fig: importance}
\end{figure*}

\subsection{Importance of each parameter}\label{sec: importance}
Random Forest (RF, ~\citep{Louppe2014}) is an ensemble machine learning algorithm that combines the predictions of multiple decision trees to improve accuracy, robustness, and generalization. This algorithm is not only a powerful predictive model but also a widely used tool for assessing feature importance. Since it builds multiple decision trees using different subsets of the data and features, it naturally provides insights into which features are most influential in making predictions. Feature importance in a RF is typically computed based on how much each feature contributes to decreasing the impurity (e.g., Gini impurity or variance) across all trees. Alternatively, it can be estimated using permutation importance, which measures the increase in prediction error when a feature’s values are randomly shuffled. These importance scores help identify the most relevant variables in a dataset, guide feature selection, and improve model interpretability. %, making Random Forest especially valuable for high-dimensional or complex data where understanding the role of each input is crucial.

We construct a RF regressor using the four selected parameters and their corresponding scores, implemented with the \texttt{scikit-learn}~\footnote{\url{https://scikit-learn.org/stable/}}. The model takes the parameter values as input features and the associated scores as target labels. We use 100 estimators in the forest, while keeping all other hyperparameters at their default settings. The resulting feature importances are shown in Figure~\ref{fig: importance}. Among the parameters, \texttt{scfind.threshold} is identified as the most influential, followed by \texttt{reliability.threshold} and \texttt{reliability.scaleKernel}, with \texttt{reliability.minSNR} contributing the least. This ranking aligns with expectations: the core detection algorithm, the S + C finder, primarily depends on \texttt{scfind.threshold}, making it the dominant factor in detection performance. In contrast, \texttt{reliability.minSNR} functions as an additional filter to suppress false positives and has limited impact when other parameters are already near optimal. 

% \subsection{Training with smaller patch}\label{sec: training smaller patch}
\subsection{Influence of training sets}\label{sec: training sets}
In Section~\ref{sec: training}, we employ three patches with size 322 (0.625$\deg^2$) as training data. Here we investigate the influence when the agent is trained on different configurations of data.
\subsubsection{Training with one patch}\label{sec: one patch}
\begin{figure*}
    \centering
    \includegraphics[width=0.49\textwidth]{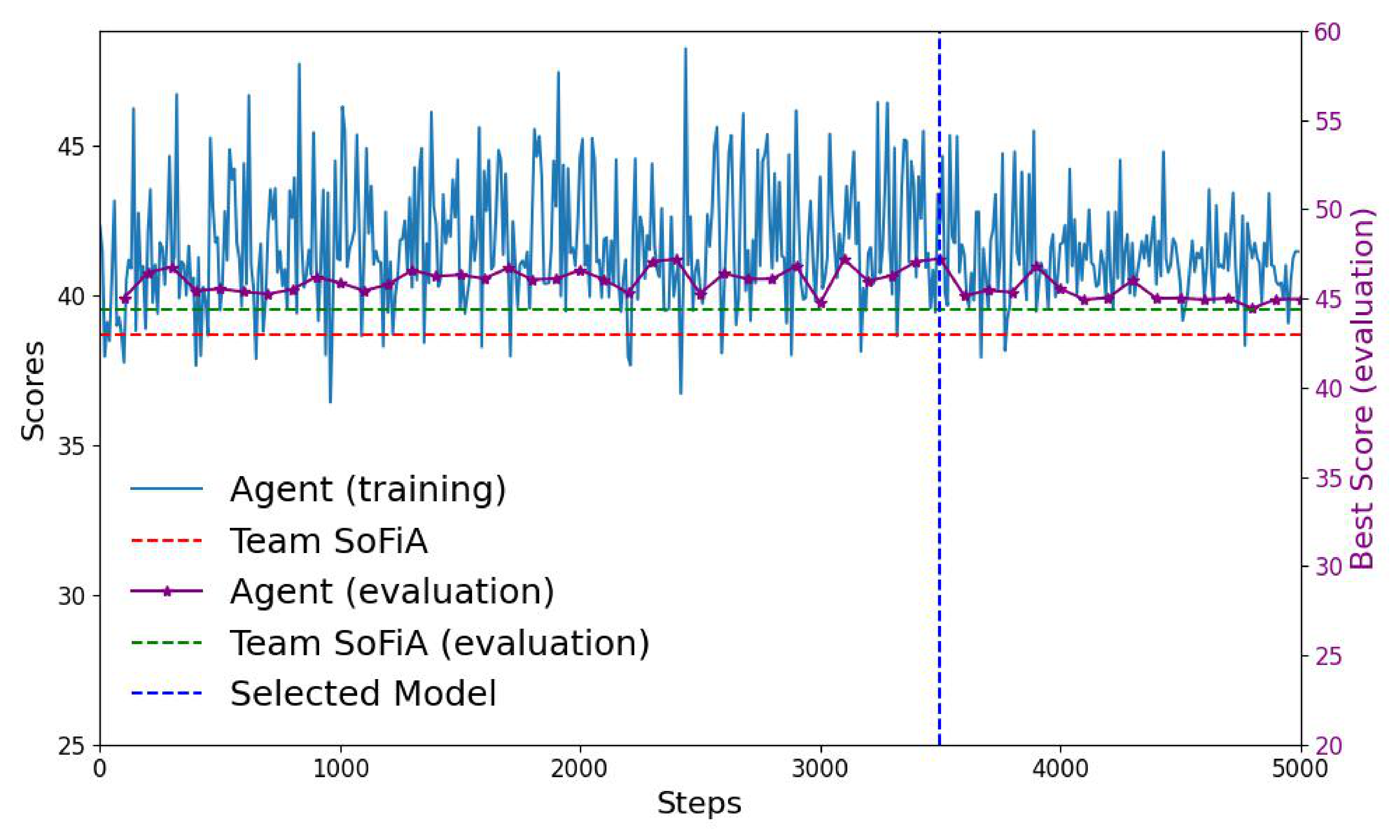}
    \includegraphics[width=0.49\textwidth]{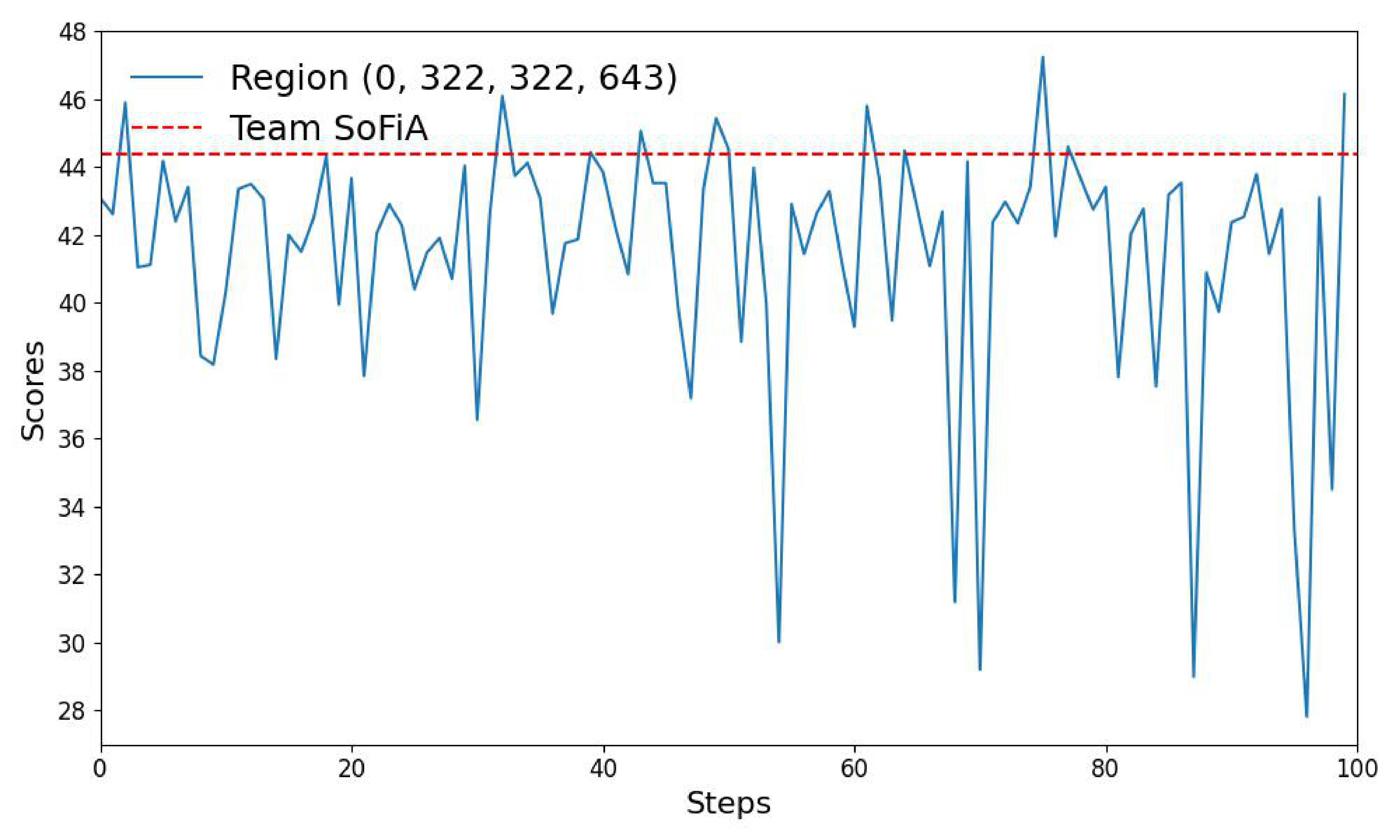}
    \caption{\textit{Left}: The training scores with respect to steps employing one single region. The testing stage is performed on the same region as mentioned in Section~\ref{sec: training}. The benchmark for training and testing patches are displayed in red and green dashed lines respectively. The best model at step 3,500 resulting highest score is selected for evaluation. \textit{Right}: The testing scores over steps for the testing patch by the best model within 100 steps. }
    \label{fig: single region}
\end{figure*}

In Section~\ref{sec: training}, we employed three patches, each covering an area of 0.0625 square degrees, from the development dataset to train the agent. Here, we assess whether training on a single region is sufficient. Specifically, we use patch (0, 322, 0, 322) and adopt the same training configuration as described in Section~\ref{sec: training}, except that the maximum number of steps is reduced to 5,000. 

The left panel of Figure~\ref{fig: single region} shows the training scores with respect to steps (blue curve), with the benchmark indicated in red dashed line. We observe that the training scores consistently exceed the benchmark, mirroring the trend for the same patch (blue line) seen in the right panel of Figure~\ref{fig: score regions}. This suggests that the parameter space in Table~\ref{tab: params} is well-optimized for this particular patch. To evaluate generalization, we conduct testing on patch (0, 322, 322, 643), the same testing patch used in Section~\ref{sec: training}. Evaluation is performed every 100 training steps, with the agent allowed up to 100 steps per evaluation to find the best parameter set. The resulting best scores are shown in purple in the same Figure. The agent consistently outperforms the benchmark at each evaluation point. The model at step 3,500 achieves the highest evaluation score and is selected as the best-performing model. The right panel of Figure~\ref{fig: single region} displays the evaluation scores over steps for this selected model. Compared to Figure~\ref{fig: scores eval model}, which corresponds to a model trained on three patches, the evaluation scores here exhibit greater fluctuations. The instability worsens as the number of steps increases, indicating that the highest score may be attributed to randomness rather than consistent performance. This variability reduces the reliability of the agent's output. Additionally, the distinct behaviors in training and evaluation stages suggest overfitting. 

Consequently, we find that training the agent on a single small region is insufficient for generalizable performance. To ensure robustness and mitigate overfitting, the agent should always be trained on large and representative sky patches. 

\subsubsection{Training with smaller patch}\label{sec: smaller patch}

\begin{figure*}
    \centering
    \includegraphics[width=\textwidth]{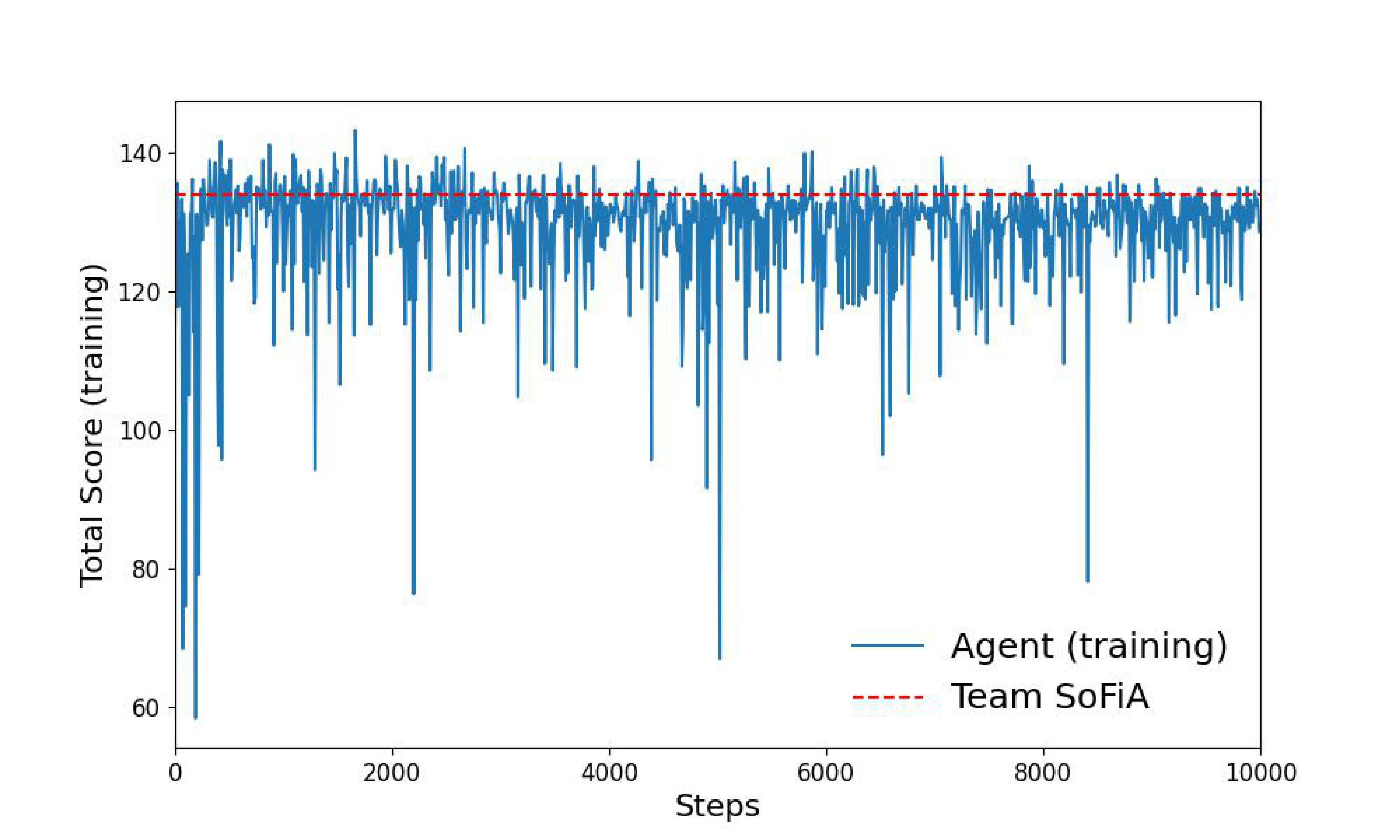}
    \caption{The aggregated training score using 12 smaller patches, each with size of 162.}
    \label{fig: small patch}
\end{figure*}

We also investigate the performance of agent trained on small patches. Here we split the sky to 16 patches, each with size $162\times162$ (0.015625 $\deg^2$). Similarly, the training data are selected based on the number of sources in individual patches, three dense, three sparse and six random ones from the rest. Note that the combined area of these training patches is the same as the ones employed in Section~\ref{sec: training}. The training scores with respect to the steps are illustrated in Figure~\ref{fig: small patch}. We notice that the convergence of the scores is difficult and the scores cannot remain steady even after 10,000 steps. The reason for this phenomenon is probably that the domination of sampling variance, since the sources are only approximately from 100 to 200 in each patch. Therefore, the number of sources also influence on the performance of the agent. 
%Therefore, we recommend that the training should be performed on large areas with abundant number of sources. 

In summary, it is recommended that training of the agent should be always performed on large and representative sky patches with abundant number of sources.

\subsubsection{Data imperfections}\label{sec: data imperfections}
{Imperfections in the data, such as artifacts, noise, or incomplete catalogs, can also affect the performance of the agent. If these imperfections can be detected and mitigated through preconditioning modules (e.g. data flagging, noise normalization, background subtraction. etc.), their impact on the resulting catalog from SoFiA-2 is negligible. Consequently, the evaluation scores and the training process of the agent remain largely unaffected. Therefore, this category of imperfection has minimal influence, provided that SoFiA-2 effectively handles it during preconditioning procedure.}

{However, if the imperfections cannot be addressed by SoFiA-2, for instance, in the case of an incomplete truth catalog, the resulting scores or rewards will be biased, causing the training and evaluation to deviate from the optimal parameters. This type of imperfection has the most significant impact on the agent's performance. To mitigate this issue, it is recommended to train the agent on a large survey area with an existing reliable catalog or on simulated data with a well-defined truth catalog.}

\subsection{Potential applications in real observation}\label{sec: potential application}

\begin{figure*}
    \centering
    \includegraphics[width=0.49\textwidth]{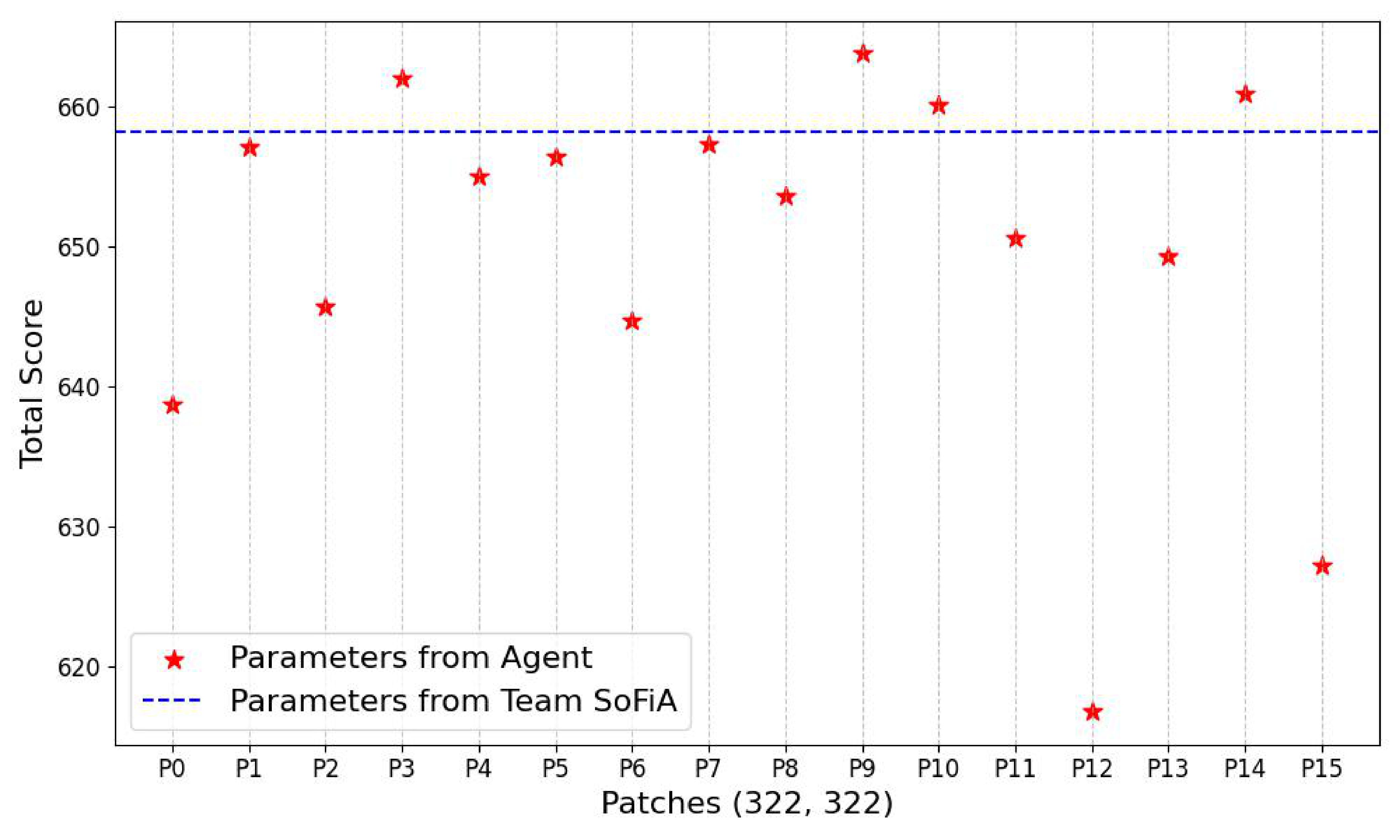}
    \includegraphics[width=0.49\textwidth]{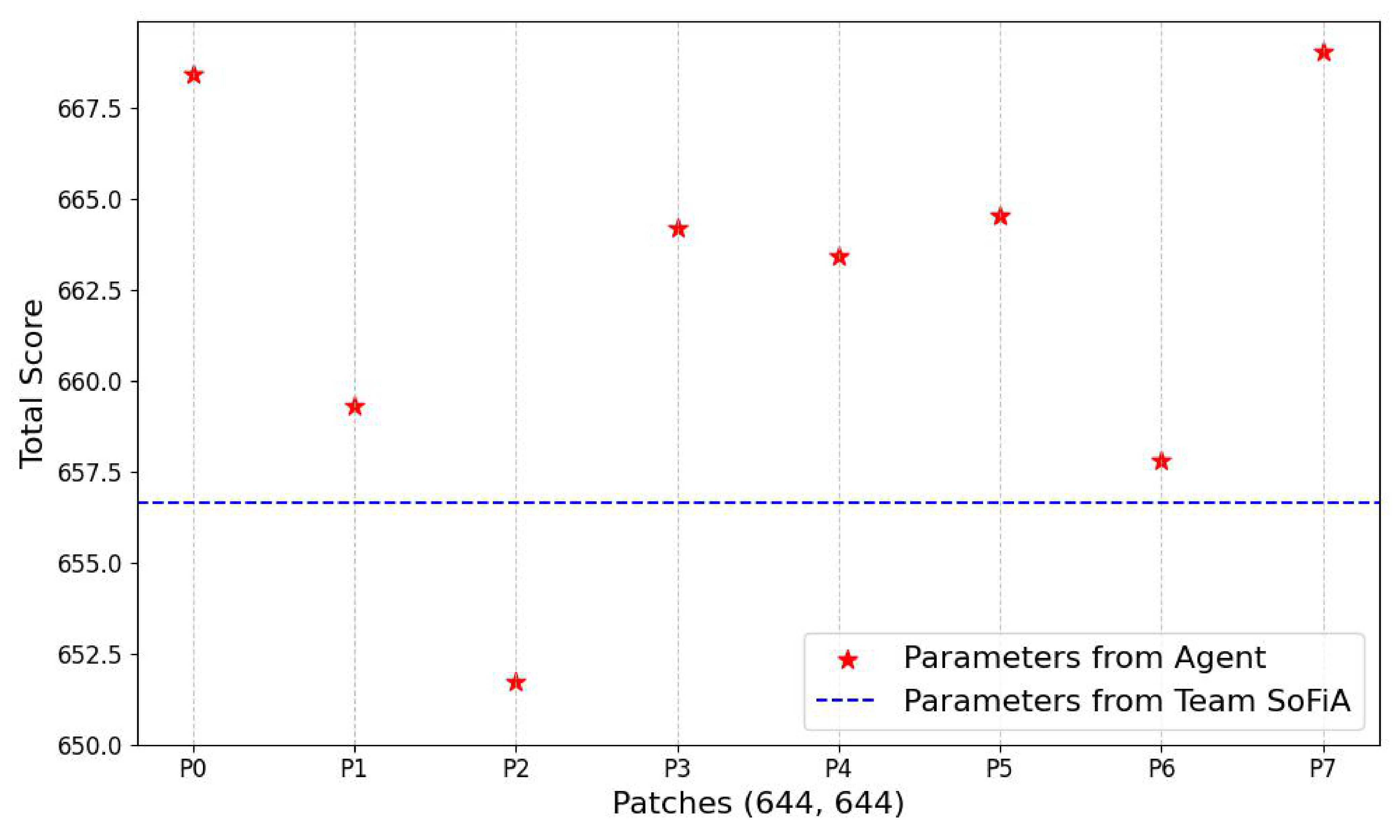}
    \caption{\textit{Left}: {The aggregated score for large development dataset employing the parameter derived by agent from patches with size of (322, 322). The blue dashed line displays the aggregated score by benchmark parameter set, and the x-axis indicates the patch IDs.} \textit{Right}: {The patches are with size of (644, 644). Note that the some patches are partially overlapped.}}
    \label{fig: total score from params}
\end{figure*}

In Section~\ref{sec: development large}, we applied the agent to each patch of area 0.0625 square degree in the large development dataset and obtained an aggregated score that surpasses the benchmark set by Team SoFiA. However, such patch-by-patch evaluation is impractical for real observations, as the agent requires interaction with environment through parameter set and the corresponding score ratio, which depends on the availability of a truth catalog for each patch. Since such catalogs are not available for real data, this limits the agent's direct deployment. 

To address this limitation, we propose a practical approach: use the agent to derive an optimal parameter set from a single representative patch, either simulated or drawn from real data with an available truth catalog, and then apply these parameters across the full observational field. To assess the effectiveness of this approach, we apply the agent on patches of spatial size 322 and 644 to find the parameter sets that yield the highest score within 100-step evaluation. These optimal parameters are then used to run SoFiA-2 across the entire dataset. To ensure consistency and restrict the computing resources, SoFiA-2 is executed on patches of same size as the one from which the parameters are derived, and then aggregate the scores across the whole dataset. The resulting total scores are shown in the left and right panels of Figure~\ref{fig: total score from params} for patch sizes 322 and 644, respectively. Benchmark scores based on parameters from Team SoFiA are calculated by aggregating scores in each patch with the same approach, and are also indicated by blue dashed lines. Note that the slight deviations of benchmark scores can be attributed to the border effect using different patch sizes. {In both panels, the x-axis shows the patch IDs. Note that for size of 644, some patches are partially overlapped.} For smaller patch size of 322, only 4 out of 16 derived parameter sets achieve total scores exceeding the benchmark. In contrast, for larger patch size of 644, 7 out of 8 parameter sets outperform the benchmark. This comparison demonstrates that larger and more representative patches are necessary to produce parameter sets with better generalization to the full dataset. 

\subsection{Limitations}\label{sec: limitations}
In Section~\ref{sec: sofia}, we focus on optimizing four continuous parameters, excluding discrete parameters such as the linking radius and minimum size used in the linker module. These discrete parameters can also be incorporated into our reinforcement learning framework, provided that practical and well-defined value ranges are specified. However, certain parameters, such as the kernel sizes in the S + C finder module, are more complex, as they are defined as lists rather than scalar values. Tuning such list-based parameters introduces additional challenges that are beyond the scope of this work and will be explored in future studies. 

The number of parameters and the size of their respective search spaces have a direct impact on the training process. Increasing either the number of parameters or the range of values typically requires more training steps for convergence. Additionally, since the current state is influenced by the actions taken in previous steps, the training must be performed sequentially. Although SoFiA-2 itself supports multi-threading execution, the sequential nature of reinforcement learning means that the training process can be computationally intensive. 

Furthermore, because the agent requires interaction with an environment that provides ground truth information, its application in real observational data is constrained by the availability of such catalogs. In practice, comprehensive truth catalogs may not available or may be too limited to support effective training. As a result, simulated observations, designed to match real instrumental and observational conditions, including relevant systematics, are required to prepare the agent for real-world deployment.

\section{Conclusions}\label{sec: conclusion}

In this work, we present an AI agent designed to optimize parameters for the software for source extraction from future large-scale sky surveys. {The agent is built using SAC algorithm, a state-of-the-art reinforcement learning method capable of efficiently handling multiple continuous parameters.} To test the feasibility and reliability of the agent, we apply it to SoFiA-2, a source-finding algorithm developed for radio observations, and use the dataset of the SKA-SDC2, an international competition designed for the development of source-finding algorithms, which includes simulated HI data cubes with realistic systematics and corresponding truth catalogs. Specifically, SoFiA-2 comprises several modules, including preconditioning, source finding, and reliability assessment, that ensure both completeness and purity of detected sources, which are governed by a set of critical parameters. Here, we focus on optimizing the detection threshold for the S + C algorithm in the core source finding module and the reliability threshold, minimum SNR, and kernel scale in the reliability module. 

The training procedure for the agent is conducted on three representative sky patches, each covering an area of 0.0625 square degrees, drawn from the SKA-SDC2 development dataset. During training, the agent automatically calls SoFiA-2 to perform source detection and compute a predefined performance metric, the SDC2 score, which accounts for both the number of matched sources and the consistency of source properties. The SAC algorithm establishes interaction with the environment through states constructed from parameter values and corresponding scores. Based on these interactions, the agent outputs actions to iteratively update the parameters in a direction that maximizes the score. 

After sufficient training, we demonstrate that the agent can identify parameter sets that outperform the benchmark provided by Team SoFiA, achieving higher scores within only 100 evaluation steps on the large development dataset. Feature importance analysis using an RF regressor confirms that the detection threshold for the S + C finder is the most influential parameter. In addition, training on larger and more representative sky patches reduces overfitting and improves generalization. To assess real-world applicability, we simulate the scenarios in which a limited patch has truth catalogs, on which we derive the parameters. Subsequently, the derived parameters are applied across the whole dataset, and the outcomes indicate that the parameters from large and representative patches can generalize well, producing higher scores compared to the benchmark.

{However, the current agent is restricted to continuous parameters, while searching for the optimal parameters in other types poses additional challenges, particularly list-based parameters of SoFiA-2.} Besides, increasing the number of parameters or expanding their value ranges significantly raises the computational cost due to the sequential nature of reinforcement learning, which requires more advanced strategies. Last but not least, the agent's dependence on ground truth limits its direct application to real observational data; consequently, high-fidelity simulated or real sky observations with validated catalogs are necessary. 

To summarize, this study demonstrates the viability of using AI agents for automatic parameter tuning in scientific software, offering a scalable solution that can be extended to other applications requiring complex and high-dimensional parameter optimization, significantly reducing the manual effort typically required, though it is still challenging in some facets.

\section{Data Availability}\label{sec: data avail}
The development and large development datasets of SKA-SDC2 are publicly available at \url{https://sdc2.skao.int/sdc2-challenge/data}, and the SDC scorer is at \url{https://gitlab.com/ska-telescope/sdc/ska-sdc}. The benchmark from Team SoFiA is at \url{https://gitlab.com/SoFiA-Admin/SKA-SDC2-SoFiA}. The relevant scripts and documentation for AI agent in this work are publicly available at \url{https://github.com/xczhou-astro/AI_agent_for_SoFiA-2}. 

\begin{acknowledgements}
XC and NL acknowledges the support of The Ministry of Science and Technology of China (No. 2020SKA0110100), The CAS Project for Young Scientists in Basic Research (No. YSBR-062). AI acknowledge postdoctoral funding from Guangdong Technion-Israel Institute of Technology, Shantou, China (ST2300022)
\end{acknowledgements}

\bibliographystyle{raa}
\bibliography{bibtex}

\appendix                  %%appendicial material is supported

\section{HI sky}\label{app: sky}
\begin{figure*}
    \centering
    \includegraphics[width=\textwidth]{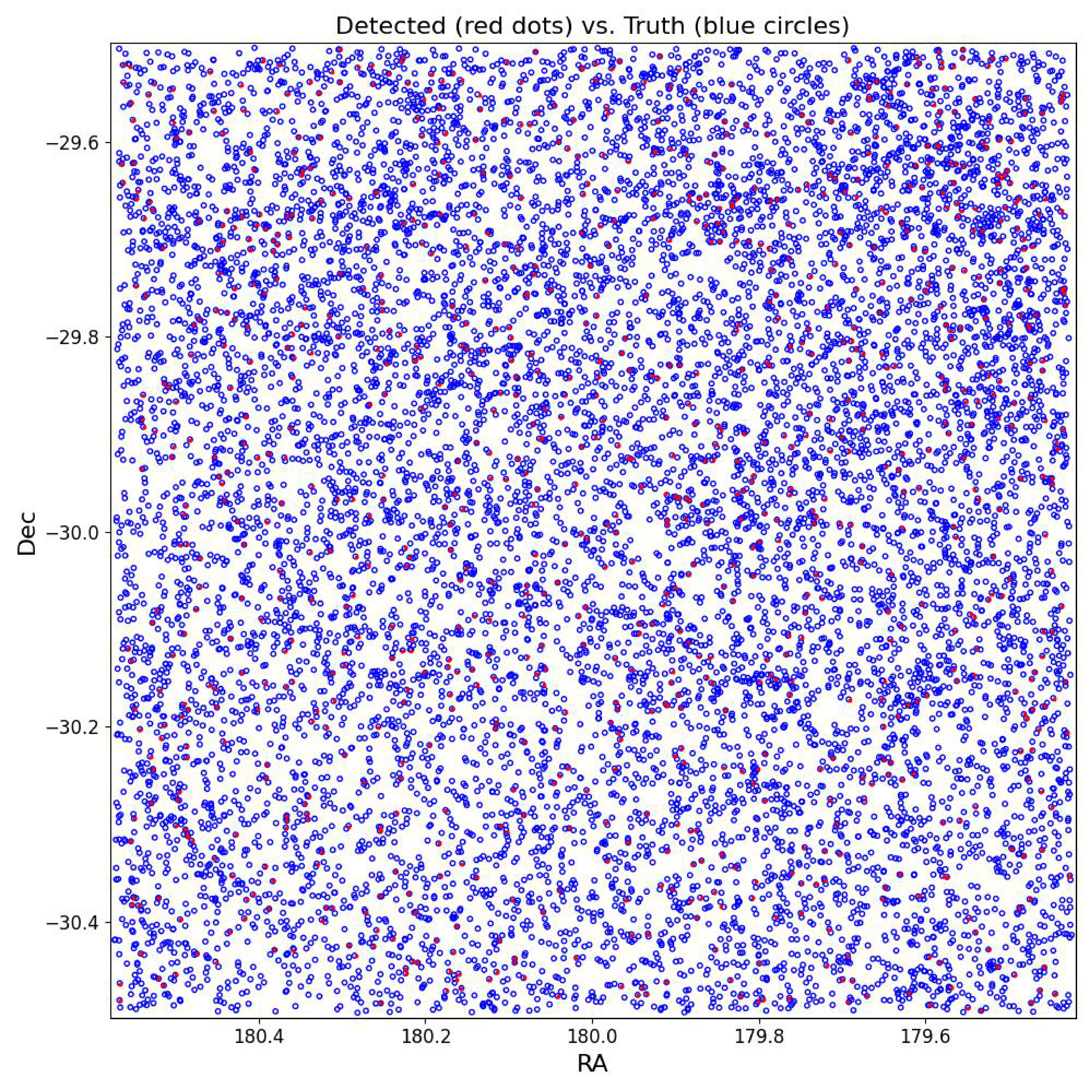}
    \caption{Sky map for large development dataset. The red dots and blue circles indicate the detected and truth sources, respectively. }
    \label{fig: hi sky}
\end{figure*}

\begin{figure*}
    \centering
    \includegraphics[width=\textwidth]{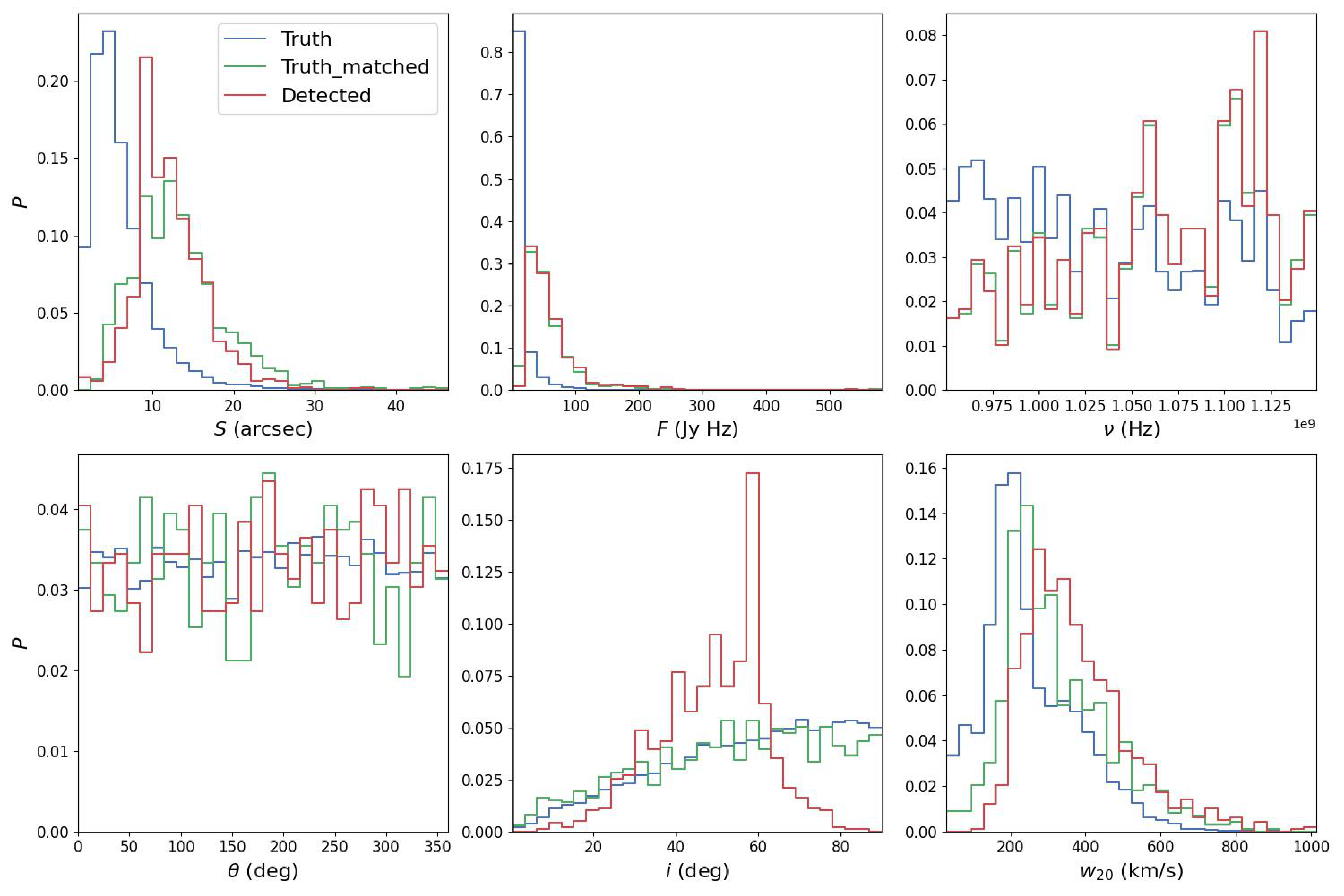}
    \caption{The distributions for source properties, including HI size ($S$), integrated line flux ($F$), central frequency ($\nu$), position angle ($\theta$), inclination angle ($i$), and line width ($w_{\rm 20}$) for full truth catalog, the subset of matched truth sources, and detected sources in large development dataset.}
    \label{fig: properties}
\end{figure*}

We present the sky coordinates of aggregated sources from both the detected and truth catalogs from large development dataset in Figure~\ref{fig: hi sky}. The detected sources are obtained using the parameter set that achieves the highest score, as determined by the agent on patch size 644 described in Section~\ref{sec: potential application}. Additionally, Figure~\ref{fig: properties} shows the distributions of key source properties, including HI size ($S$), integrated line flux ($F$), central frequency ($\nu$), position angle ($\theta$), inclination angle ($i$), and line width ($w_{\rm 20}$), for the full truth catalog, the subset of matched truth sources, and the detected sources. The observed deviations in HI size and inclination for the detected sources arise from the use of assumed mean values for unresolved sources, following the recommendations in the conversion scripts provided by Team SoFiA~\footnote{\url{https://gitlab.com/SoFiA-Admin/SKA-SDC2-SoFiA/-/blob/master/scripts/physical_parameter_conversion.py?ref_type=heads}}.

\label{lastpage}

\end{document}